\documentclass[letterpaper]{article} % DO NOT CHANGE THIS
\usepackage{aaai2026}  % DO NOT CHANGE THIS
\usepackage{times}  % DO NOT CHANGE THIS
\usepackage{helvet}  % DO NOT CHANGE THIS
\usepackage{courier}  % DO NOT CHANGE THIS
\usepackage[hyphens]{url}  % DO NOT CHANGE THIS
\usepackage{graphicx} % DO NOT CHANGE THIS
\usepackage{natbib}  % DO NOT CHANGE THIS AND DO NOT ADD ANY OPTIONS TO IT
\usepackage{caption} % DO NOT CHANGE THIS AND DO NOT ADD ANY OPTIONS TO IT
\usepackage{algorithm}
\usepackage{algorithmic}

\usepackage{tikz}
\usepackage{amsmath} 
\usepackage{booktabs} 
\usepackage{xcolor} 
\usepackage{amssymb}
\usepackage{xcolor}
\usepackage{array} 
\usepackage{subcaption}
\usepackage{multirow}
\usepackage{booktabs} % For professional looking tables
\usepackage{textcomp}
\usepackage{newfloat}
\usepackage{listings}
\usepackage{tabularx}
\usepackage{booktabs}
\DeclareCaptionStyle{ruled}{labelfont=normalfont,labelsep=colon,strut=off} % DO NOT CHANGE THIS
\floatstyle{ruled}
\newfloat{listing}{tb}{lst}{}
\floatname{listing}{Listing}
\title{MPA: Multimodal Prototype Augmentation for Few-Shot Learning}
\author{
    Liwen Wu\textsuperscript{\rm 1,2},
    Wei Wang\textsuperscript{\rm 1},
    Lei Zhao\textsuperscript{\rm 3},
    Zhan Gao\textsuperscript{\rm 3},
    Qika Lin\textsuperscript{\rm 4},
    Shaowen Yao\textsuperscript{\rm 1},
    Zuozhu Liu\textsuperscript{\rm 5},
    Bin Pu\textsuperscript{\rm 3}\footnote{Corresponding author.}
}
\affiliations{
    \textsuperscript{\rm 1}Yunnan University, Kunming, China\\
    \textsuperscript{\rm 2}Engineering Research Center of Cyberspace, Kunming, China\\
    \textsuperscript{\rm 3}Hunan University, Changsha, China\\
    \textsuperscript{\rm 4}National University of Singapore, Singapore\\
    \textsuperscript{\rm 5}Zhejiang University, Hangzhou, China\\
    
    \{lwwu,yaosw\}@ynu.edu.cn, \{zhaolei,gzhan,pubin\}@hnu.edu.cn, weiwang@stu.ynu.edu.cn, zuozhuliu@intl.zju.edu.cn
}

\usepackage{bibentry}
\begin{document}

\maketitle

\begin{abstract}
Recently, few-shot learning (FSL) has become a popular task that aims to recognize new classes from only a few labeled examples and has been widely applied in fields such as natural science, remote sensing, and medical images.
However, most existing methods focus only on the visual modality and compute prototypes directly from raw support images, which lack comprehensive and rich multimodal information.
To address these limitations, we propose a novel \textbf{\underline{\emph{M}}}ultimodal \textbf{\underline{\emph{P}}}rototype \textbf{\underline{\emph{A}}}ugmentation FSL framework called \textbf{MPA}, including LLM-based Multi-Variant Semantic Enhancement (LMSE), Hierarchical Multi-View Augmentation (HMA), and an Adaptive Uncertain Class Absorber (AUCA). 
LMSE leverages large language models to generate diverse paraphrased category descriptions, enriching the support set with additional semantic cues. 
HMA exploits both natural and multi-view augmentations to enhance feature diversity (e.g., changes in viewing distance, camera angles, and lighting conditions). 
AUCA models uncertainty by introducing uncertain classes via interpolation and Gaussian sampling, effectively absorbing uncertain samples. 
Extensive experiments on four single-domain and six cross-domain FSL benchmarks demonstrate that MPA achieves superior performance compared to existing state-of-the-art methods across most settings. Notably, MPA surpasses the second-best method by 12.29\% and 24.56\% in the single-domain and cross-domain setting, respectively, in the 5-way 1-shot setting.

\end{abstract}

\begin{links}
    \link{Code}{https://github.com/ww36user/MPA}
\end{links}

\begin{figure}[!t]
\centering
\includegraphics[width=\linewidth]{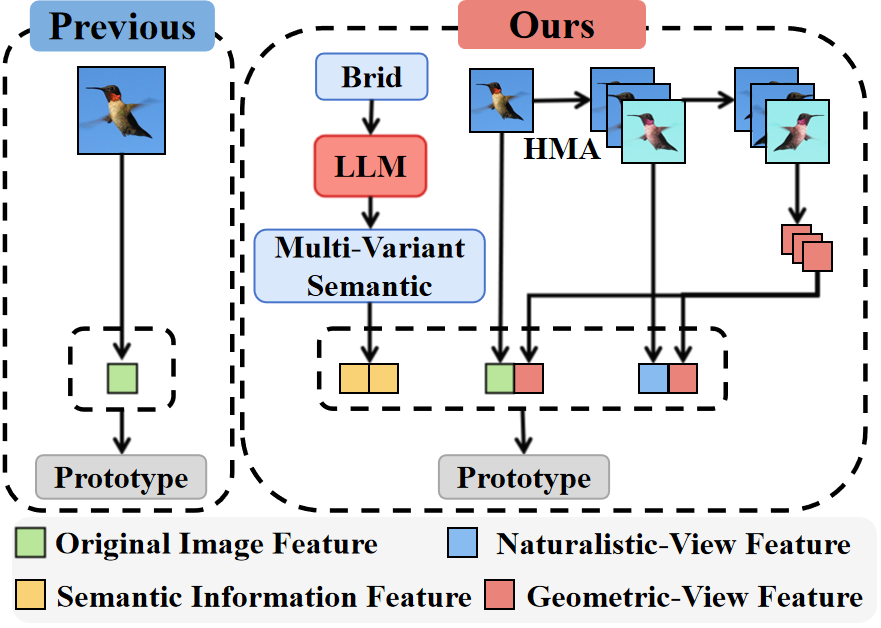}
\caption{Unlike existing methods that typically focus on original image feature-based prototype, MPA integrates multimodal feature-based prototype (i.e., LLM-based multi-variant semantic features and hierarchical Multi-View features), significantly improving the model’s generalization performance and robustness across diverse tasks.}
\label{proto}  
\end{figure}

\section{Introduction}
In recent years, few-shot learning (FSL) has attracted widespread attention for its ability to learn efficiently with limited labeled data \citep{song2024self, liu2025envisioning,zou2024flatten,zhu2025anatomical,li2024tkr}, and has been widely applied in many fields such as image recognition \citep{su2025prototypeformer,cai2024little}, crop disease diagnosis\citep{zhang2024exploring}, and remote sensing \citep{ma2025reconstruction}.
Prototype-based metric methods have been a popular direction in FSL due to their simplicity, efficiency, and generalization ability \citep{zhang2021prototype}.
These methods typically focus on single-modal, single-view image features to construct prototypes and classify based on the distance between query samples and prototypes.

For example, FSL-PRS utilizes a self-attention mechanism to jointly leverage support sets and query sets, dynamically adjusting visual prototypes to enhance classification discrimination capabilities \citep{zhao2024few}. However, due to its focus solely on the visual modality, its performance remains limited.

Similarly, TEFSL proposes a few-shot learning framework that enhances spectral-spatial embedding, improving classification performance through attention feature embedding and graph iterative prototype optimization. However, it still focuses solely on the visual modality, neglecting the importance of semantic information \citep{xi2025transductive,11151642}.
A key limitation remains: \emph{existing methods often overlook the potential of multimodal prototypes and multi-view representations in few-shot scenarios.} 

As shown in Figure~\ref{proto}, single-modal/single-view-based methods may fail to adequately capture the diverse characteristics of limited-support set image samples. 

For example, consider a bird image where different species share very similar visual appearances—similar colors, patterns, and shapes—making it difficult for single-modal methods to distinguish between them. However, incorporating semantic information such as species descriptions (“small, with a sharp, curved beak and a bold stripe across its wings”) can provide crucial clues that complement the visual data, enabling more accurate recognition. 
\textbf{\emph{(1) This highlights the limitation of relying solely on visual modality and the advantage of leveraging semantic modality, since a bird’s representation is inherently multi-dimensional and cannot be fully captured through visual information alone.}} Especially by leveraging large language models to fully harness the semantic representation capabilities of language-based multimodal information, the semantic richness of samples can be significantly enhanced.

Similarly, while few-shot images offer limited expressive information on their own, leveraging multiple views can fully capture their representations. 
\textbf{\emph{(2) In other words, natural variations in viewing distance, camera angles, and lighting conditions provide diverse perspectives that enhance feature diversity.}}
In short, incorporating multimodal and multi-view augmentations can enrich sample representations and significantly enhance both discrimination and generalization.

Based on the above analysis, we propose a novel multimodal prototype augmentation FSL framework called MPA, which integrates LLM-based multi-variant semantic enhancement (LMSE), hierarchical multi-view augmentation (HMA), and adaptive uncertain class absorption (AUCA), effectively overcoming these limitations and providing a more comprehensive and robust solution to improving prototype representativeness in few-shot learning.
Specifically, LMSE is designed to capture high-quality semantic features during prototype construction, thereby breaking the dependence on image features. Then, HMA is proposed to leverage natural variations and geometric views to generate diverse representations, effectively enriching the support set and enhancing feature diversity. Finally, AUCA combines interpolation with normal distribution dynamic sampling to generate an adaptive uncertain class for absorbing uncertain samples and mitigating inter-class interference. Throughout the process, we utilized the CLIP model to extract multimodal features from multi-variant semantic information, and multi-view visual information. Finally, we trained a logistic regression classifier using these features, whose effectiveness was thoroughly validated across multiple evaluation tasks. The major contributions of this study are outlined as:

\begin{itemize}
    \item We introduce an LLM-based Multi-Variant Semantic Enhancement module that generates high-quality semantic features from class names and descriptions, reducing reliance on image-only features in prototype construction.
    \item We design a Hierarchical Multi-View Augmentation module that leverages natural data augmentations and geometric views to generate diverse feature representations, thereby significantly enriching feature diversity and enhancing prototype robustness.
    \item We propose an Adaptive Uncertain Class Absorber that generates uncertain classes via interpolation and normal distribution sampling, effectively absorbing uncertain samples and reducing inter-class interference.
    \item Extensive experiments on both single-domain and cross-domain FSL datasets, validate the effectiveness of MPA in addressing the challenges of limited data and complex labeling in healthcare, achieving outstanding performance across diverse settings.
\end{itemize}

\begin{figure*}[!t]
\centering
\includegraphics[width=\linewidth]{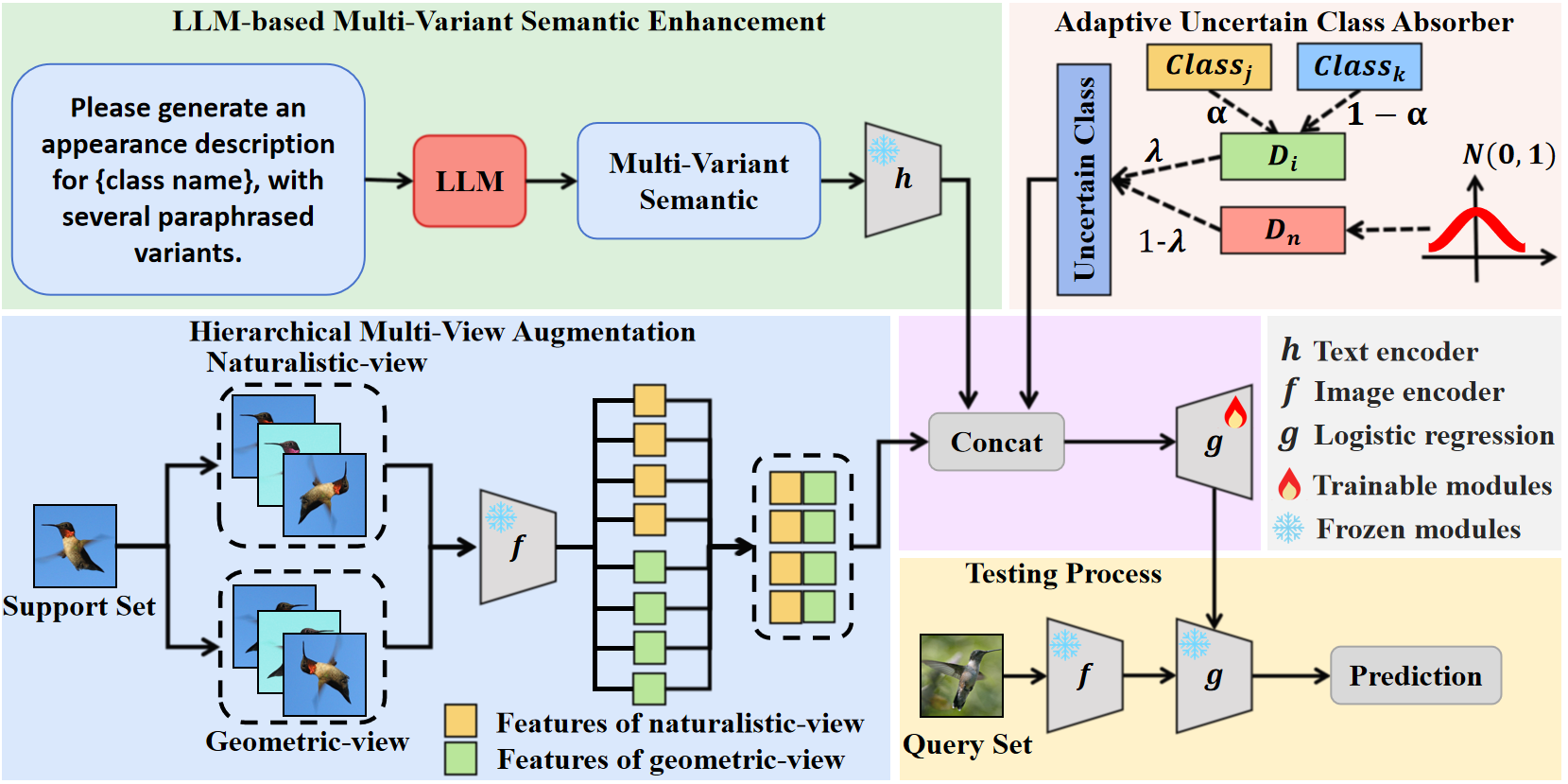}
\caption{Overview of MPA. Our framework includes three components: LLM-based Multi-Variant Semantic Enhancement (LMSE), Hierarchical Multi-View Augmentation (HMA), and Adaptive Uncertain Class Absorber (AUCA). LMSE produces high-quality semantic features by leveraging LLM, reducing reliance on visual inputs for prototype construction. HMA improves data diversity and feature robustness through multi-view and feature-level augmentations. AUCA mitigates sample bias by interpolating between prototypes and sampling from a normal distribution, where the interpolation weight \( \lambda \) is adaptively determined based on prototype differences. Finally, logistic regression is applied to optimized features for classification.}

\label{frame_image}  
\end{figure*}

\section{Related Work}
\subsection{Few-shot Learning} 
Few-shot learning aims to generalize to novel classes using only a limited number of labeled samples. Existing approaches can be grouped into metric-based \citep{su2025prototypeformer, li2022ranking,zhang2024exploring}, optimization-based \citep{ma2025reconstruction,song2024self,zou2024flatten}, transfer-based \citep{fu2023styleadv, liu2025envisioning}, and data augmentation-based methods. Metric-based methods aim to construct an embedding space where samples from the same class are closely clustered, while those from different classes are well separated. For instance, PrototypeFormer \citep{su2025prototypeformer} introduces a transformer-based Prototype Extraction Module that leverages learnable prototype tokens to capture highly discriminative class representations. Additionally, it employs a Prototype Contrastive Loss to improve the robustness and generalization of prototypes, achieving state-of-the-art performance on few-shot learning benchmarks. Similarly, SEVPro \citep{cai2024little} improves prototype quality by mapping semantic cues to boost class separation, making it easier to distinguish similar categories. This method significantly improves performance across various prototype-based FSL tasks.

\subsection{Cross-Domain Few-shot Learning}
Despite its progress, FSL models often suffer performance degradation when applied to classes with significant domain shifts. To tackle domain shift, various approaches have been proposed \citep{zhou2023revisiting,wang2023diffusion,wang2025fedfsl,cai2024little,zhang2024simple,liu2025envisioning,liu2025making}. For example, Zhou et al. \citep{zhou2023revisiting} proposed a local-global distillation framework, representing the first effort to combine both local and global distillation strategies to tackle domain shifts and improve cross-domain generalization. Similarly, SPM \citep{song2024self} introduces semantic information as prompts to adaptively guide the feature extractor, enhancing class-specific representations and boosting generalization under domain shift. However, previous prototype-based methods often lack semantic richness and multi-view representations, which limits their ability to fully capture feature distributions and reduces their robustness in few-shot scenarios. However, although SPC \citep{11151642} focuses on local semantic cues to refine models, it still ignores the importance of multi-view diversity and rich textual semantics. The method mainly relies on intra-image spatial relations, lacking complementary visual views and textual guidance. As a result, its representations are biased toward limited visual contexts, reducing adaptability and generalization to unseen domains or ambiguous classes.

\section{Methodology}
\subsection{Overview}
As shown in Figure \ref{frame_image}, we propose a multimodal support set enhancement framework to improve prototype quality and generalization in few-shot learning.
Specifically, we first introduce an LLM-based multi-variant semantic enhancement method, which provides high-quality class-level semantic features to complement visual features and alleviate reliance on limited image data. 
Then, a hierarchical multi-view augmentation strategy is employed, combining naturally transformed naturalistic-views and geometric-views to enrich the diversity of the support set and enhance the expressiveness of feature representations. 
To further enhance robustness, we introduce an adaptive uncertain class absorber that adaptively constructs uncertain classes via interpolation and normal distribution sampling, helping to absorb uncertain samples.
Together, these components form a unified approach that strengthens prototype representations, mitigates noise interference, and significantly improves model generalization in FSL scenarios.

\subsection{LLM-based Multi-Variant Semantic Enhancement}

To enrich class representations, we generate diverse semantic variants using an language model, encode them, and integrate them into the support set.
Unlike images with noisy backgrounds, semantic descriptions focus on key attributes, thus improving feature diversity.
The support image features are extracted using the CLIP image encoder \citep{radford2021learning}, and can be expressed as:
\begin{equation}
F_{\mathrm{i}} = f\bigl(\{I_{n}\}_{n=1}^{N}\bigr) = \bigl\{f(I_{1}), f(I_{2}), \ldots, f(I_{N})\bigr\},
\end{equation}
where \( \{I_{n}\}_{n=1}^{N} \) represents the set of all \(N\) images in the support set, and \( f(\cdot) \) is the CLIP image encoder function applied to each image individually.

To obtain richer and more diverse semantic representations, we use a large language model (LLM, e.g., GPT-4.0) to generate multi-variant semantics for each class with the prompt: \textit{``Please generate an appearance description for \{class name\}, with four paraphrased variants.''} Specifically, we define a generation function $\mathcal{G}_{\text{LLM}}$ that takes a class label $c$ as input and produces a set of semantically enriched information $\{ t_m \}_{m=1}^M$ (including the original description and four paraphrased variants) to capture diverse and complementary semantic features. These semantic representations are then projected into the embedding space using the CLIP text encoder $h(\cdot)$, expressed as:

\begin{equation}
F_{\mathrm{t}} = h\left( \mathcal{G}_{\text{LLM}}(c) \right),
\label{eq:text}
\end{equation}
where \( \mathcal{G}_{\text{LLM}}(c) = \{ t_m \}_{m=1}^M \) represents the set of $M$ multi-variant semantic informations generated for class \( c \), and \( h(\cdot) \) represents the CLIP text encoder that projects the textual variants into a shared semantic space. 
As a result, \( F_{\mathrm{t}} \in \mathbb{R}^d \) is the resulting semantic embedding that captures the enriched textual information. 
By incorporating LLM-generated variants, we introduce contextual diversity and latent knowledge into the textual representation, which significantly enhances generalization under few-shot conditions and alleviates the limitations of fixed or manually written class names.

\subsection{Hierarchical Multi-View Augmentation}

The hierarchical multi-view augmentation module aims to enrich the support set and improve generalization in few-shot scenarios by introducing diverse visual perspectives.
We first apply transformations such as central cropping, rotation, and color perturbation to simulate natural variations.
Visually diverse yet label-consistent, these augmented samples help reduce dependence on scarce data.

Specifically, given a support image \( I \), the naturalistic-view is generated as:
\begin{equation}
\text{\( I \)}_a = \{ \tau_n(I) \mid \tau_n \in \mathcal{T} \},
\label{eq:aug_set}
\end{equation}
where $\mathcal{T}$ represents the set of individual augmentation functions applied independently to $I$, and $\mathcal{I}_a$ is the resulting set of augmented images generated by applying each transformation $\tau_n$ separately. In HMA, to enhance the diversity of visual inputs, we apply hierarchical augmentations including central crops (120, 170, 200 pixels), rotations (45\textdegree, 90\textdegree, 180\textdegree, 270\textdegree, 315\textdegree), and color jitter (brightness = 0.5, contrast = 0.5, saturation = 0.5, hue = 0.2). The corresponding features are then extracted by an image encoder:
\begin{equation}
F_{\mathrm{a}} = f\bigl(\{I_{a}^{(m)}\}_{m=1}^{M}\bigr) = \bigl\{f(I_{a}^{(1)}), f(I_{a}^{(2)}), \ldots, f(I_{a}^{(M)})\bigr\},
\label{eq:ssa2}
\end{equation}
where \( \{I_{a}^{(m)}\}_{m=1}^{M} \) represents the set of \( M \) naturalistic views generated from a raw support image \( I \), and \( f(\cdot) \) is the CLIP image encoder function applied to each naturalistic-view individually, resulting in a set of feature vectors \( F_{\mathrm{a}} \in \mathbb{R}^{M \times d} \).

In parallel, complementary views are generated via horizontal reflection of support images, and their features are combined with the originals to enrich representation and boost few-shot generalization.

\subsection{Adaptive Uncertain Class Absorber} 
To enhance model generalizability, we introduce an adaptive uncertain class absorber by leveraging interpolation and sampling from a normal distribution. The interpolation-based data can be formulated as:

\begin{equation}
\text{D}_{i} = 
\begin{bmatrix}
\alpha & 1 - \alpha
\end{bmatrix}
\cdot
\begin{bmatrix}
\text{F}_j \\
\text{F}_k
\end{bmatrix},
\label{eq:adcg1} 
\end{equation}
where the interpolation coefficient $\alpha$ ranges from 0.2 to 0.8 in our experiments, and $\text{F}_j$, $\text{F}_k$ denote features from different classes.
Next, the normal distribution is expressed as:
\begin{equation}
\text{D}_{\text{n}} \sim \mathcal{N}(0,1),
\end{equation}
where \( \text{Data}_{\text{n}} \in \mathbb{R}^{1 \times 768} \). 
Then, the cosine similarity \(S_{j,k}\) between the prototypes of class \(j\) and class \(k\) is computed as:
\begin{equation}
  S_{j,k} = \frac{\text{C}_j \cdot \text{C}_k}{\|\text{C}_j\| \|\text{C}_k\|}, \quad j \leq k,
\end{equation}
where \(\|\cdot\|\) represents the Euclidean norm of a vector.
Based on the cosine similarity, construct a symmetric matrix \(\mathbf{S}\), where the lower triangular part is set to 0:
\begin{equation}
  \mathbf{S} = \begin{bmatrix}
    S_{1,1} & S_{1,2} & \cdots & S_{1,N} \\
    0       & S_{2,2} & \cdots & S_{2,N} \\
    \vdots  & \vdots  & \ddots & \vdots  \\
    0       & 0       & \cdots & S_{N,N}
  \end{bmatrix} ,
\end{equation}
where \(N\) is the total number of classes. 
Normalize the matrix \(\mathbf{S}\) globally to map the cosine similarity to the range \([0, 1]\):
\begin{equation}
S'_{j,k} = \arg\min_{\lambda \in \mathbb{R}} \left\| S_{j,k} - \left( (1 - \beta) S_{\text{min}} + \beta S_{\text{max}} \right) \right\|^2 ,
\end{equation}
where \(S_{\text{min}}\) and \(S_{\text{max}}\) are the minimum and maximum values of the matrix \(\mathbf{S}\), respectively. 

Finally, compute average of the normalized similarities for all class pairs and convert it into a difference measure:

\begin{equation}
\lambda = 1 - \frac{2}{\binom{C}{2}} \sum_{j=1}^{C-1} \sum_{k=j+1}^{C} S'_{j,k},
\end{equation}
where \(\binom{C}{2}\) is the combination number. 
As a result, the uncertain class data can be expressed as:
\begin{multline}
\mathbb{E}[\text{D}_{\text{u}}] = (1-\lambda) \cdot \text{D}_{\text{n}} + \lambda \cdot \text{D}_{\text{i}}, \\
\text{D}_{\text{u}} \sim
\begin{cases}
\text{D}_{\text{n}}, & P(\text{D}_{\text{u}} = \text{D}_{\text{n}}) = 1 - \lambda \\
\text{D}_{\text{i}}, & P(\text{D}_{\text{u}} = \text{D}_{\text{i}}) = \lambda
\end{cases} \mathrel{\phantom{=}},
\label{eq:adcg2}
\end{multline}
where \(\lambda\) is a dynamic factor determined by the inter-prototype differences within each batch. \(\text{D}_{\text{u}}\), \(\text{D}_{\text{n}}\), and \(\text{D}_{\text{i}}\) denote uncertain-class, Gaussian-sampled, and interpolated data, respectively.

In cross-domain scenarios, where features are more clustered and harder to distinguish, $\lambda$ is typically smaller; in contrast, it tends to be larger in single-domain datasets with more separable features. MPA injects diversity via interpolation and Gaussian sampling to foster robust decision boundaries and flexible overlap with real classes.

\begin{table*}[h]
    \centering
    \begin{tabular}{c c c | c c c c | c}
        \toprule
        \textbf{Method} & \textbf{Marker} & \textbf{PT} & \textbf{miniImageNet} & \textbf{tieredImageNet} & \textbf{CIFAR-FS} & \textbf{FC100} & \textbf{Average} \\
        \toprule
        Meta-UAFS  & IJCAI-21 & \textbf{$\times$} & 64.22 & 69.13 & 74.08 & 41.99 & 62.36 \\ 
        HCTransformers  & CVPR-22 & \textbf{$\times$} & 74.62 & 79.57 & 78.88 & 48.15 & 70.31 \\ 
        
        CPEA  & ICCV-23 & \textbf{$\times$} & 71.97 & 76.93 & 77.82 & 47.24 & 68.49 \\ 
        Diff-ResNet  & TPAMI-24 & \textbf{$\times$} & 73.47 & 79.74 & \textbf{-} & \textbf{-} & \textbf{-} \\ 
        FedFSL-CFRD   & AAAI-25 & \textbf{$\times$}  & 56.66 & 60.52 & 63.00 & \textbf{-} & \textbf{-} \\
        \midrule
        SP-CLIP  & CVPR-23 & \textbf{$\checkmark$} & 72.31 & 78.03 & 82.18 & 48.53 & 70.26 \\ 
        
        CAML  & ICLR-24 & \textbf{$\checkmark$} & 96.20 & {95.40} & 70.80 & \textbf{-} & \textbf{-} \\ 
        SEVPro  & IJCAI-24 & \textbf{$\checkmark$} & 71.81 & 72.77 & 80.36 & \textbf{-} & \textbf{-} \\ 
        SemFew-Trans  & CVPR-24 & \textbf{$\checkmark$} & 78.94 & 82.37 & 84.34 & 54.27 & 74.98 \\ 
        SPM  & AAAI-24 & \textbf{$\checkmark$} & 93.70 & 88.79 & 82.40 & \underline{68.35} & \underline{83.31} \\ 
        
        ECER-FSL      & AAAI-25 & \textbf{$\checkmark$}  & 81.14 & 81.81 & 86.01 & 57.34 & 76.58 \\
        
        MLVLM         & AAAI-25 & \textbf{$\checkmark$}  & \underline{98.24} & \underline{98.06} & \underline{95.02} & \textbf{-} & \textbf{-} \\  \midrule
        \textbf{MPA (Ours)} & \textbf{-} & \textbf{$\checkmark$} & \textbf{98.87} & \textbf{98.57} & \textbf{97.47} & \textbf{87.47} & \textbf{95.60} \\ 
        \bottomrule
    \end{tabular}
    \caption{Performance comparison of state-of-the-art methods on 5-way 1-shot single-domain datasets. \textbf{PT} represents a public pre-trained model. The best results are in \textbf{bold}, and the second-best are \underline{underlined}.}
    \label{tab:sota1_single}
\end{table*}

\begin{table*}[h]
    \centering    
    \begin{tabular}{c c c | c c c c c c | c}
        \toprule
        \textbf{Method} & \textbf{Mark }& \textbf{PT} & \textbf{CUB} & \textbf{Cars} & \textbf{Places} & \textbf{Plantae} & \textbf{EuroSAT} & \textbf{CropDisease} & \textbf{Average} \\ 
        \midrule
        TPN+ATA  & IJCAI-21 & \textbf{$\times$} & 51.89 & 38.07 & 57.26 & 40.75  & 70.84 & 82.47 & 56.88 \\
        RDC  & CVPR-22 & \textbf{$\times$} & 50.09 & 39.94 & 61.17 & 41.30  & 70.51 & {85.79} & 58.13 \\
        LDP-net & CVPR-23 & \textbf{$\times$} & 55.94 & 37.44 & 62.21 & 41.04  & 73.25 & 81.24 & 58.52 \\
        
        TPN+FAP & IJCAI-24 & \textbf{$\times$} & 50.56 & 34.39 & 57.34 & 37.44  & 62.62 & 76.11 & 53.08 \\
        PRM-Net & NIPS-24 & \textbf{$\times$} & 59.48 & 38.86 & 62.90 & 44.06  & 69.56 & 84.01 & 59.81 \\
        
        \midrule
        StyleAdv-FT  & CVPR-23 & \textbf{$\checkmark$} & 84.01 & 40.48 & 72.64 & 55.52 & 74.93 & 84.11 & 68.62 \\
        FloR  & CVPR-24 & \textbf{$\checkmark$} & 55.94 & 40.01 & 61.27 & 41.70  & 71.38 & {86.30} & 59.43 \\
        SPM & AAAI-24 & \textbf{$\checkmark$} & 84.39 & {41.71} & 72.35 & 53.85  & 74.97 & 84.43 & 68.62 \\
        SVasP & AAAI-25 & \textbf{$\checkmark$} & {85.56} & 40.51 & \underline{75.93} & \underline{56.25} & {75.51} & 83.98 & \underline{69.62} \\
        % GNN+DKM  & Arxiv-25 & 51.09 & 38.61 & 60.57 & 40.47 & \textbf{-} & \textbf{-} & \textbf{-} & \textbf{-} \\
        DAMIM  & AAAI-25 & \textbf{$\checkmark$} & \textbf{-} & \textbf{-} & \textbf{-} & \textbf{-} & \underline{77.23} & \underline{86.74} & \textbf{-} \\
        ECER-FSL & AAAI-25 & \textbf{$\checkmark$} & \textbf{-} & \textbf{-} & \textbf{-} & \textbf{-}  & 74.13 & 82.13 & \textbf{-} \\
        MLVLM  & AAAI-25 & \textbf{$\checkmark$} & \underline{96.64} & \textbf{99.74} & \textbf{-}  & \textbf{-} & \textbf{-} & \textbf{-} & \textbf{-} \\ \midrule
        \textbf{MPA (Ours)} & \textbf{-} & \textbf{$\checkmark$} & \textbf{98.95} & \underline{98.51} & \textbf{93.55} & \textbf{91.73} & \textbf{87.05} & \textbf{95.28} & \textbf{94.18} \\
        \bottomrule
    \end{tabular}
    \caption{Comparison with state-of-the-art methods on 5-way 1-shot classification across cross-domain datasets.}
    \label{tab:sota1_cross}
\end{table*}

\section{Experiments}

\subsection{Datasets and Evaluation}
We conducted extensive experiments on ten datasets, including four single-domain and six cross-domain datasets. The publicly available datasets used in our experiments are divided into two groups. The single-domain datasets consist of miniImageNet \citep{vinyals2016matching}, tieredImageNet \citep{tieredImagenet}, CIFAR-FS \citep{CIFARFS}, and FC100 \citep{CIFAR100}. The cross-domain datasets include fine-grained classification benchmarks CUB \citep{Welinder2010CUB} and Cars \citep{krause2013Cars}, which present subtle inter-class differences, as well as more general domain datasets such as Places \citep{wang2020Places}, Plantae \citep{van2018Plantae}, EuroSAT \citep{helber2019EuroSAT}, and CropDisease \citep{mohanty2016CropDisease}.

\subsection{Baseline Methods}

We compare our model with a broad range of state-of-the-art baselines to validate the effectiveness of MPA.
For cross-domain FSL tasks, we evaluate against:
TPN+ATA \citep{ijcai2021-149}, RDC \citep{li2022ranking}, LDP-Net \citep{zhou2023revisiting}, TPN+FAP \citep{zhang2024exploring}, PRM-Net \citep{zhou2024meta}, StyleAdv-FT \citep{fu2023styleadv}, FloR \citep{zou2024flatten}, SVasP \citep{li2025svasp}, and DAMIM \citep{ma2025reconstruction}. For standard FSL tasks, we additionally include comparisons with:
Meta-UAFS \citep{zhang2021uncertainty}, HCTransformers \citep{he2022attribute}, CPEA \citep{hao2023class}, Diff-ResNet \citep{wang2023diffusion}, FedFSL-CFRD \citep{wang2025fedfsl}, SP-CLIP \citep{Chen2023Semantic}, CAML \citep{fifty2023context}, SEVPro \citep{cai2024little}, SemFew-Trans \citep{zhang2024simple} and DAMIM \citep{ma2025reconstruction}. Notably, SPM \citep{song2024self}, ECER-FSL \citep{liu2025envisioning}, and MLVLM \citep{liu2025making} are evaluated under both single-domain and cross-domain FSL settings.

\subsection{Implementation Details}
Our model is implemented using the PyTorch framework, employing the CLIP model with the ViT-L/14 backbone as the feature extractor. In each epoch, we randomly sample 100 episodes from the target domain. For each task, the support set consists of 5 classes, with each class containing either 5 or 1 sample, while the query set includes 15 images per class. All experiments were conducted on a workstation equipped with an NVIDIA RTX 4090 GPU, Intel(R) Xeon(R) Silver 4310 CPU, and 32GB RAM.

\begin{table*}[h]
    \centering
    \begin{tabular}{c c c | c c c c | c}
        \toprule
        \textbf{Method} & \textbf{Mark} & \textbf{PT} & \textbf{miniImageNet} & \textbf{tieredImageNet} & \textbf{CIFAR-FS} & \textbf{FC100} & \textbf{Average} \\
        \midrule
        Meta-UAFS         & IJCAI-21 & \textbf{$\times$} & 79.99 & 84.33 & 85.92 & 57.43 & 76.92 \\
        HCTransformers    & CVPR-22  & \textbf{$\times$} & 89.19 & 91.72 & 90.50 & 66.42 & 84.46 \\
        CPEA              & CVPR-23  & \textbf{$\times$} & 87.26 & 90.12 & 88.98 & 65.02 & 82.85 \\
        Diff-ResNet       & TPAMI-24 & \textbf{$\times$} & 83.86 & 87.10 & -    & -    & -    \\
        FedFSL-CFRD       & AAAI-25  & \textbf{$\times$} & 72.15 & 78.95 & 74.58 & -    & -    \\
        \midrule
        SP-CLIP           & CVPR-23  & \textbf{$\checkmark$} & 83.42 & 88.55 & 88.24 & 61.55 & 80.44 \\
        CAML              & ICLR-24  & \textbf{$\checkmark$} & \underline{98.60} & \underline{98.10} & 85.50 & -    & -    \\
        SEVPro       & IJCAI-24 & \textbf{$\checkmark$} & 78.88 & 84.04 & 86.12 & -    & -    \\
        SemFew-Trans      & CVPR-24  & \textbf{$\checkmark$} & 86.49 & 89.89 & 89.11 & 65.02 & 82.63 \\
        SPM               & AAAI-24  & \textbf{$\checkmark$} & 98.30 & 96.20 & \underline{93.10} & \underline{83.78} & \underline{92.85} \\
        \midrule
        \textbf{MPA (Ours)} & \textbf{-} & \textbf{$\checkmark$} & \textbf{99.07} & \textbf{98.79} & \textbf{97.64} & \textbf{90.57} & \textbf{96.52} \\
        \bottomrule
    \end{tabular}
    \caption{Performance comparison of state-of-the-art methods on 5-way 5-shot single-domain datasets.}
    \label{tab:sota5_single}
\end{table*}

\begin{table*}[!t]
    \centering
    \begin{tabular}{c c c | c c c c c c | c}
        \toprule
        \textbf{Method} & \textbf{Mark} & \textbf{PT} & \textbf{CUB} & \textbf{Cars} & \textbf{Places} & \textbf{Plantae} & \textbf{EuroSAT} & \textbf{CropDisease} & \textbf{Average} \\ 
        \midrule
        TPN+ATA      & IJCAI-21   & \textbf{$\times$} & 70.14 & 55.23 & 73.87 & 59.02 & 85.47 & 93.56 & 72.88 \\
        RDC          & CVPR-22    & \textbf{$\times$} & 67.23 & 53.49 & 74.91 & 57.47 & 84.29 & 93.30 & 71.78 \\
        LDP-Net      & CVPR-23    & \textbf{$\times$} & 73.34 & 53.06 & 75.47 & 59.64 & 82.01 & 89.40 & 72.15 \\
        TPN+FAP      & IJCAI-24   & \textbf{$\times$} & 64.17 & 47.38 & 72.05 & 53.58 & 80.24 & 88.34 & 67.63 \\
        PRM-Net      & NeurIPS-24 & \textbf{$\times$} & 76.68 & 55.44 & 76.98 & 63.08 & 83.22 & 93.09 & 74.75 \\
        \midrule
        StyleAdv-FT  & CVPR-23    & \textbf{$\checkmark$} & 95.82 & 66.02 & 88.33 & 78.01 & 90.12 & 95.99 & 85.72 \\
        FloR         & CVPR-24    & \textbf{$\checkmark$} & 74.06 & 57.98 & 74.25 & 61.70 & 83.76 & 93.60 & 74.23 \\
        SPM          & AAAI-24    & \textbf{$\checkmark$} & 95.95 & 62.89 & 88.79 & 74.52 & 89.72 & 96.11 & 84.66 \\
        SVasP        & AAAI-25   & \textbf{$\checkmark$} & \underline{95.95} & \underline{66.47} & \underline{89.19} & \underline{78.67} & 90.55 & 96.17 & \underline{86.17} \\
        DAMIM        & AAAI-25    & \textbf{$\checkmark$} & -     & -     & -     & -     & \underline{91.08} & \underline{96.49} & - \\
        \midrule
        \textbf{MPA (Ours)} & \textbf{-} & \textbf{$\checkmark$} & \textbf{99.32} & \textbf{99.63} & \textbf{93.37} & \textbf{95.37} & \textbf{91.77} & \textbf{98.21} & \textbf{96.28} \\
        \bottomrule
    \end{tabular}
    \caption{Comparison with state-of-the-art methods on 5-way 5-shot classification across cross-domain datasets.}
    \label{tab:sota5_cross}
\end{table*}

\begin{table}[h]
\centering
\begin{tabular}{c | c c | c c}
\toprule
\multirow{2}{*}{\textbf{Method}} & \multicolumn{2}{c|}{\textbf{CUB}} & \multicolumn{2}{c}{\textbf{Cars}} \\
& \textbf{1-shot} & \textbf{5-shot} & \textbf{1-shot} & \textbf{5-shot} \\
\midrule
TPN+ATA      & 51.89 & 70.14 & 38.07 & 55.23 \\
RDC          & 50.09 & 67.23 & 39.94 & 53.49 \\
LDP-net      & 55.94 & 73.34 & 37.44 & 53.06 \\
TPN+FAP      & 50.56 & 64.17 & 34.39 & 47.38 \\
PRM-Net      & 59.48 & 76.68 & 38.86 & 55.44 \\
StyleAdv-FT  & 84.01 & 95.82 & 40.48 & 66.02 \\
FloR         & 55.94 & 74.06 & 40.01 & 57.98 \\
SPM          & 84.39 & 95.95 & 41.71 & 62.89 \\
SVasP        & 85.56 & \underline{95.95} & 40.51 & \underline{66.47} \\
MLVLM        & \underline{96.64} & -     & \textbf{99.74} & - \\
\textbf{MPA (Ours)} & \textbf{98.95} & \textbf{99.32} & \underline{98.51} & \textbf{99.63} \\
\bottomrule
\end{tabular}
\caption{Comparison results on fine-grained datasets (CUB and Cars) under 5-way 1-shot and 5-shot settings.}
\label{tab:finegrained_combined}
\end{table}

\subsection{Comparison with State-of-the-arts} 

\subsubsection{Overall Results.} 
As shown in Tables \ref{tab:sota1_single}, \ref{tab:sota1_cross}, \ref{tab:sota5_single}, and \ref{tab:sota5_cross}, MPA exhibits outstanding classification performance and broad applicability across all experimental conditions. Notably, our method performs effectively not only in natural scenarios (e.g.,  miniImageNet and CIFAR-FS datasets) but also achieves strong results in fine-grained tasks (CUB and Cars datasets). Moreover, MPA demonstrates robust capabilities in single-domain few-shot learning (miniImageNet, tieredImageNet, CIFAR-FS and FC100 datasets), cross-domain few-shot learning (EuroSAT and CropDisease), and fine-grained recognition tasks (Cars, CUB, Places and Plantae datasets). These results highlight MPA’s potential for practical use across a wide range of real-world scenarios.\\
\textbf{Results on single-domain FSL datasets.} 
As shown in Table \ref{tab:sota1_single}, under the 5-way 1-shot setting, the average classification accuracy of MPA surpasses the second-best method by 12.29\%. 
Specifically, MPA improves the average accuracy by 5.17\%, 9.78\%, 15.07\%, and 19.12\% compared to the second-best method on miniImageNet, tieredImageNet, CIFAR-FS and FC100 datasets, respectively.
As shown in Table \ref{tab:sota5_single}, under the 5-way 5-shot setting, the average classification accuracy of MPA surpasses the second-best method (SPM) by 3.67\%. 
Specifically, MPA improves the average classification accuracy by 0.77\%, 2.59\%, 4.54\%, and 6.79\% compared to the second-best method on miniImageNet, tieredImageNet, CIFAR-FS and FC100 datasets, respectively.\\

\noindent\textbf{Results on cross-domain FSL datasets.} 
As shown in Table \ref{tab:sota1_cross}, MPA achieves superior performance on all six datasets under the 5-way 1-shot setting. Specifically, MPA achieves the best performance on CUB, Places, Plantae, EuroSAT, and CropDisease datasets and ranks second-best on Cars, with an average improvement of 24.56\%.
As shown in Table \ref{tab:sota5_cross}, under the 5-way 5-shot setting, MPA achieves the best performance across all six datasets. 
Specifically, MPA improves the average accuracy by 3.37\%, 33.16\%, 4.18\%, 16.7\%, 1.22\% and 2.04\% compared to the second-best method on CUB, Cars, Places, Plantae, EuroSAT and CropDisease datasets, respectively.\\
\noindent\textbf{Results on fine-grained FSL datasets.} 
As shown in Table~\ref{tab:finegrained_combined}, MPA performs exceptionally well on the CUB and Cars datasets in the 5-way 1-shot and 5-way 5-shot settings, achieving the best results in three metrics and ranking second in the other.
Specifically, in the 5-way 5-shot classification task on the Cars dataset, MPA achieves a 33.16\% improvement in accuracy compared to the second-place model.  
These results demonstrate that MPA possesses significant advantages and strong generalization capabilities in low-sample fine-grained classification tasks.  \\

\subsection{Ablation Study}
\subsubsection{Effect of MPA Components.} 
As presented in Table~\ref{tab:mpa}, to comprehensively evaluate the contributions of each component in our MPA framework, we conduct an ablation study across three datasets.
Without introducing HMA and AUCA, using LMSE alone can significantly improve performance on EuroSAT and CIFAR-FS. 
After introducing HMA, performance is further improved on Places and CIFAR-FS, indicating that multi-view augmentation helps model generalization.
Finally, when the three modules are used together, the best performance is achieved on all datasets, showing the effectiveness and complementarity of each module.

\begin{table}[H]
\centering
\setlength{\tabcolsep}{4.5pt} % 控制列间距
\begin{tabular}{ccc|ccc}
\toprule
\textbf{LMSE} & \textbf{HMA} & \textbf{AUCA} &
\textbf{EuroSAT} & \textbf{Places} & \textbf{CIFAR-FS} \\
\midrule
$\times$ & $\times$ & $\times$ & 76.41 & 87.24 & 93.69  \\
$\checkmark$ & $\times$ & $\times$ & 83.03 & \underline{93.43} & 95.36  \\
$\times$ & $\checkmark$ & $\times$ & 79.44 & 84.71 & 94.17  \\
$\checkmark$ & $\checkmark$ & $\times$ & \underline{85.69} & 92.64 & \underline{96.32} \\
$\checkmark$ & $\checkmark$ & $\checkmark$ & \textbf{87.05} & \textbf{93.55} & \textbf{97.47} \\ 
\bottomrule
\end{tabular}
\caption{Ablation study on the effects of the three components across three datasets in 5-way 1-shot setting.}
\label{tab:mpa}
\end{table}

\subsubsection{Effect of Backbone.}
To evaluate the generalizability of MPA under the challenging 5-way 1-shot setting, we conduct experiments on multiple CLIP backbone variants, including ViT-L/14, ViT-B/32, ViT-B/16, and ResNet101.
As shown in Table~\ref{tab:backbone_comparison}, MPA consistently outperforms the baseline across all backbones and datasets, demonstrating strong robustness and adaptability in few-shot scenarios.

\begin{table}[h]
\centering

\begin{tabular}{c c |cccc}
\toprule

\textbf{Method} & \textbf{Backbone} & \textbf{EuroSAT} & \textbf{Places} & \textbf{FC100} \\
% \midrule
% Second-best & - & 77.23 & 75.93  & 68.35 \\
\midrule
Baseline & ViT-L/14 & 76.41 & 87.24  & 77.18 \\
\textbf{Ours} & ViT-L/14 & \textbf{87.05} & \textbf{93.55}  & \textbf{87.47} \\
\midrule
Baseline & ViT-B/32 & 65.47 & 82.32  & 60.20 \\
\textbf{Ours} & ViT-B/32 & \textbf{74.15} & \textbf{91.08}  & \textbf{75.95} \\
\midrule
Baseline & ViT-B/16 & 71.19 & 83.92  & 67.95 \\
\textbf{Ours} & ViT-B/16 & \textbf{78.92} & \textbf{92.00} & \textbf{79.73} \\
\midrule
Baseline & ResNet101 & 57.88 & 78.71  & 46.40 \\
\textbf{Ours} & ResNet101 & \textbf{65.99} & \textbf{83.99} & \textbf{51.61} \\
\bottomrule
\end{tabular}
\caption{Performance comparison across different CLIP backbones on three datasets under the 5-way 1-shot setting.}

\label{tab:backbone_comparison}
\end{table}

\subsubsection{Extended Evaluation and Analysis.}
To further validate the generality and robustness of our approach, we provide several extended experiments and analyses. 
Specifically, Table \ref{tab:llmgpt} examines the impact of different large language models on semantic enhancement, and GPT-4.0 achieves the best accuracy. 
{\textit{Appendix. A2}\textsuperscript{†}}
evaluates the effectiveness of AUCA under extreme conditions with uncertain samples. 
A statistical analysis of the dynamic factor $\lambda$ is presented in {\textit{Appendix. A3}\textsuperscript{†}} to demonstrate the adaptability of AUCA across various domains. 
UMAP visualizations of MPA features on five target-domain datasets are provided in {\textit{Appendix. A4}\textsuperscript{†}}, 
showing that MPA clearly separates category boundaries compared with the baseline. 
Additional feature visualizations and analysis are presented in {\textit{Appendix. A5}\textsuperscript{†}}.

These findings consistently support the effectiveness of our method in learning transferable representations and addressing diverse data distributions.

\begin{table}[H]
    \centering
    \begin{tabular}{c|cc}
        \toprule
        \textbf{LLM}  & \textbf{tieredImageNet} &  \textbf{CIFAR-FS}   \\
        \midrule
        GPT-3.5  & 97.88 & 97.20  \\
        \textbf{GPT-4.0}  & \textbf{98.57} & \textbf{97.47}  \\
        DeepSeek-V3  & 97.79 & 97.01  \\
        DeepSeek-R1  & 97.79 & 97.25   \\
        ChatGLM-3  & 97.99 & 97.19   \\
        Claude-4  & {98.17} & {97.39}   \\
        Gemini-2.5  & 98.17 & 96.53   \\
        InternVL3  & 98.16 & 97.15  \\
        Qwen2.5  & 98.15 & 97.36   \\
        ERNIE X1  & 98.03 & 97.32  \\
    
        \bottomrule
    \end{tabular}
    \caption{Performance comparison of LMSE using different LLMs, evaluated on 5-way 1-shot tasks across two datasets.}
    \label{tab:llmgpt}
\end{table}

\section{Conclusion} 
In this paper, we propose MPA, a novel multimodal few-shot learning framework called MPA, which includes LLM-based Multi-Variant Semantic Enhancement (LMSE), Hierarchical Multi-View Augmentation (HMA), and Adaptive Uncertain Class Absorber (AUCA) for improving generalization and stability. LMSE is designed to generate high-quality semantic features, reducing reliance on image data. HMA enriches support representations by combining natural augmentations and geometric views. AUCA dynamically generates adaptive uncertain classes through interpolation and normal distribution sampling to address uncertain samples and reduce class overlap. Experiments on four single-domain datasets and six cross-domain datasets show state-of-the-art performance, demonstrating strong generalization and effectiveness in various FSL scenarios.

\clearpage

\section{Acknowledgments}
This study is supported by the Yunnan Fundamental Research Projects (Nos. 202301AU070194 and 202501AT070233), the Yunnan Province Special Project (No. 202403AP140021), the Practical Innovation Project of Postgraduate Students in the Professional Degree of Yunnan University (No. ZC-24248950), the Yunnan Province expert workstations (No. 202305AF150078), and the National Natural Science Foundation of China (Nos. 62162067 and 62562061).

% \section{Acknowledgements}
% This study is supported by the Yunnan Fundamental Research Projects (Nos. 202301AU070194 and 202501AT070233), and the Yunnan Province Special Project (No. 202403AP140021).
\bibliography{aaai2026}

% \clearpage
%调用另一个文件
% \input{ReproducibilityChecklist/LaTeX/ReproducibilityChecklist}

\clearpage

\section{Appendix}
\subsection{A1. Training Procedure}

In the few-shot image recognition setting, we first extract features from each image along with its naturalistic-views and geometric-views using the CLIP image encoder, capturing both natural variations and spatial transformations. Multi-variant semantic information of class names is simultaneously obtained via the CLIP text encoder. To handle outliers and enhance representation, uncertain classes are dynamically generated to absorb uncertain samples. A logistic regression classifier is then trained on the enriched support set features, and finally, the classifier is used to predict labels for the query images. The detailed description and algorithm of our MPA framework are provided in Algorithm 1.
% Algorithm
\begin{algorithm}[H]
\caption{The procedure of MPA to train the classifier.}
\label{alg:algorithm}
\textbf{Input}: Target domain dataset $D_t$, CLIP image encoder $f_\theta$, CLIP text encoder $h_{\theta}$.\\

\textbf{Output}: Trained logistic regression parameters ${\theta_{\text{logic}}}$. 

\begin{algorithmic}[1] %[1] enables line numbers

% \STATE Let $t=0$.
\WHILE{training}
\STATE Randomly sample a few-shot task $T$ from $D_t$.
\STATE Enrich the support set representation via Hierarchical Multi-View Augmentation, as formulated in Eqs.~\eqref{eq:aug_set} and~\eqref{eq:ssa2}.

\STATE Extract multi-variant semantic representations by prompting the LLM with support set class names, as formulated in Eq.~\eqref{eq:text}.

\STATE Project the derived semantic embeddings into the visual space via $h_{\theta}$; adapt dimensionality through vector duplication and incorporate into the support set.

\STATE Synthesize uncertain-class samples leveraging the generative mechanism described in Eqs.~\eqref{eq:adcg1} and~\eqref{eq:adcg2}, and integrate them into the support set.

\STATE Perform supervised optimization over the enriched support set using a logistic regression classifier.

\STATE Update the logistic regression parameters $\theta_{\text{logic}}$ accordingly.

\ENDWHILE
\STATE \textbf{return} ${\theta_{\text{logic}}}$ .
\end{algorithmic}
\end{algorithm}

\subsection{A2. AUCA Under Extreme Conditions}
To evaluate AUCA's robustness, we introduce artificial uncertainty under the 5-way 5-shot setting. As shown in Table \ref{tab:adcg2}, AUCA consistently improves performance, achieving a 2.97\% gain on CUB and a 2.39\% average improvement on the target domain.
\begin{table}[H]
    \centering
    \begin{tabular}{c|cccc}
        \toprule
        \textbf{AUCA}  & \textbf{CUB} & \textbf{Cars} &  \textbf{CIFAR-FS} & \textbf{FC100}  \\
        \midrule
        $\times$  & 82.75 & 84.96  & 73.89 & 60.29  \\
        $\checkmark$  & \textbf{85.72} & \textbf{87.44}  & \textbf{75.96} & \textbf{62.31}  \\
        \bottomrule
    \end{tabular}
    \caption{Average classification accuracy across target domain datasets, including those with artificially added uncertain samples, under the 5-way 5-shot setting.}
    \label{tab:adcg2}
\end{table}

\begin{table}[H]
    \centering

    \begin{tabular}{@{}c  c | c c@{}} 
        \toprule
        \textbf{Dataset}  & \textbf{Recognition task} & \textbf{Mean} & \textbf{Variance} \\
        \toprule
        
        CUB & Fine-grained bird & 0.3150 & 0.0011 \\
        Cars & Fine-grained car & 0.3646 & 0.0014 \\
        Places & Scene & 0.4727 & 0.0018 \\
        Plantae & Plantae & 0.3236 & 0.0012 \\
        % \toprule
        EuroSAT & Satellite images & 0.2244 & 0.0013 \\
        CropDisease & Agricultural disease & 0.2208 & 0.0014 \\
        \toprule
        miniImageNet &General few-shot & 0.3947 & 0.0015 \\
        tieredImageNet &General few-shot  & 0.4257 & 0.0013 \\
        CIFAR-FS &General few-shot  & 0.3002 & 0.0009 \\
        FC100 & General few-shot  & 0.2779 & 0.0011 \\
        \bottomrule
    \end{tabular}
    \caption{Mean values and variances of the dynamic factor \(\lambda\) across various datasets from different domains.}
    \label{tab:lambda}
\end{table}

\begin{figure}[H]
\centering
\includegraphics[width=\linewidth]{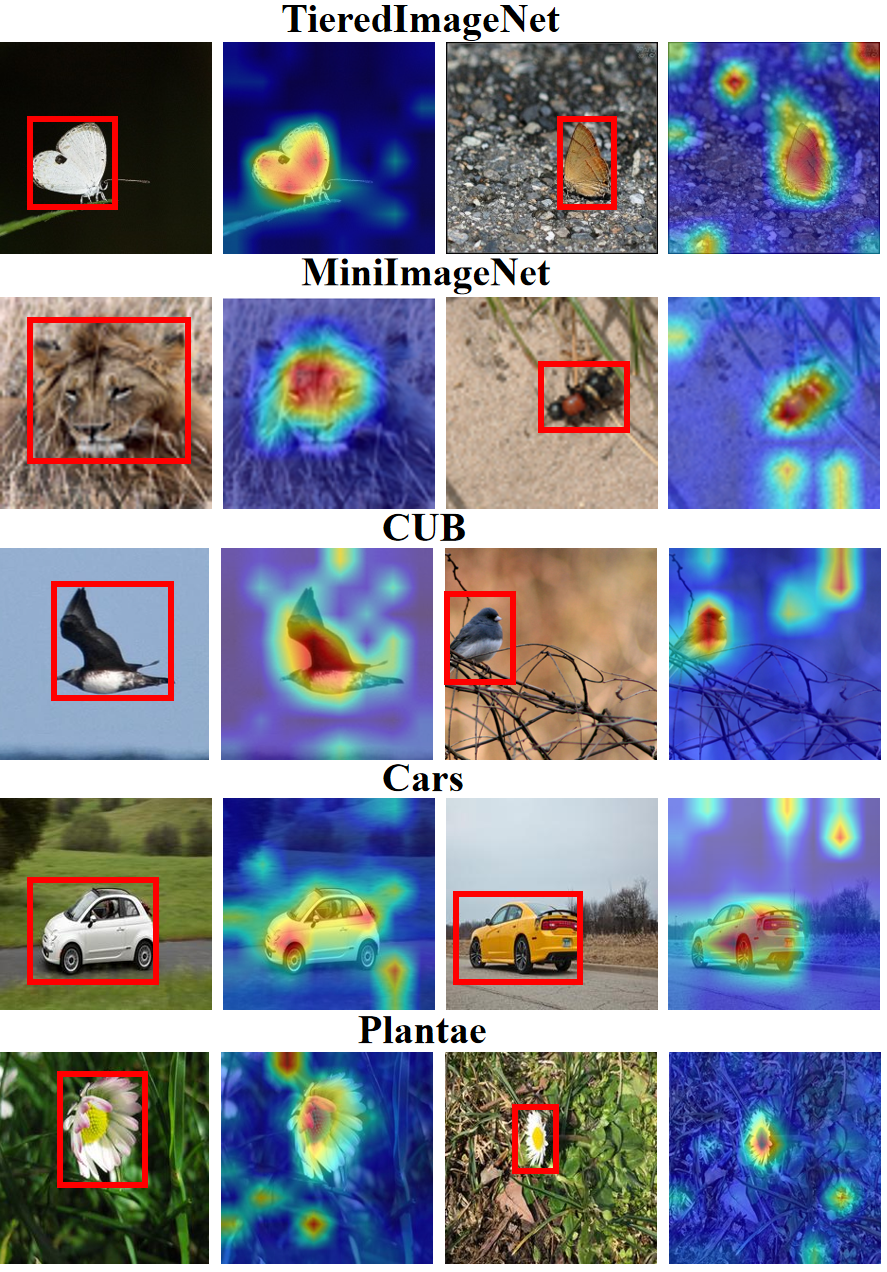}
\caption{Feature visualization of MPA on public datasets.}
\label{cam}  
\end{figure}

\begin{figure*}[!t]
\centering
\includegraphics[width=0.69\linewidth]{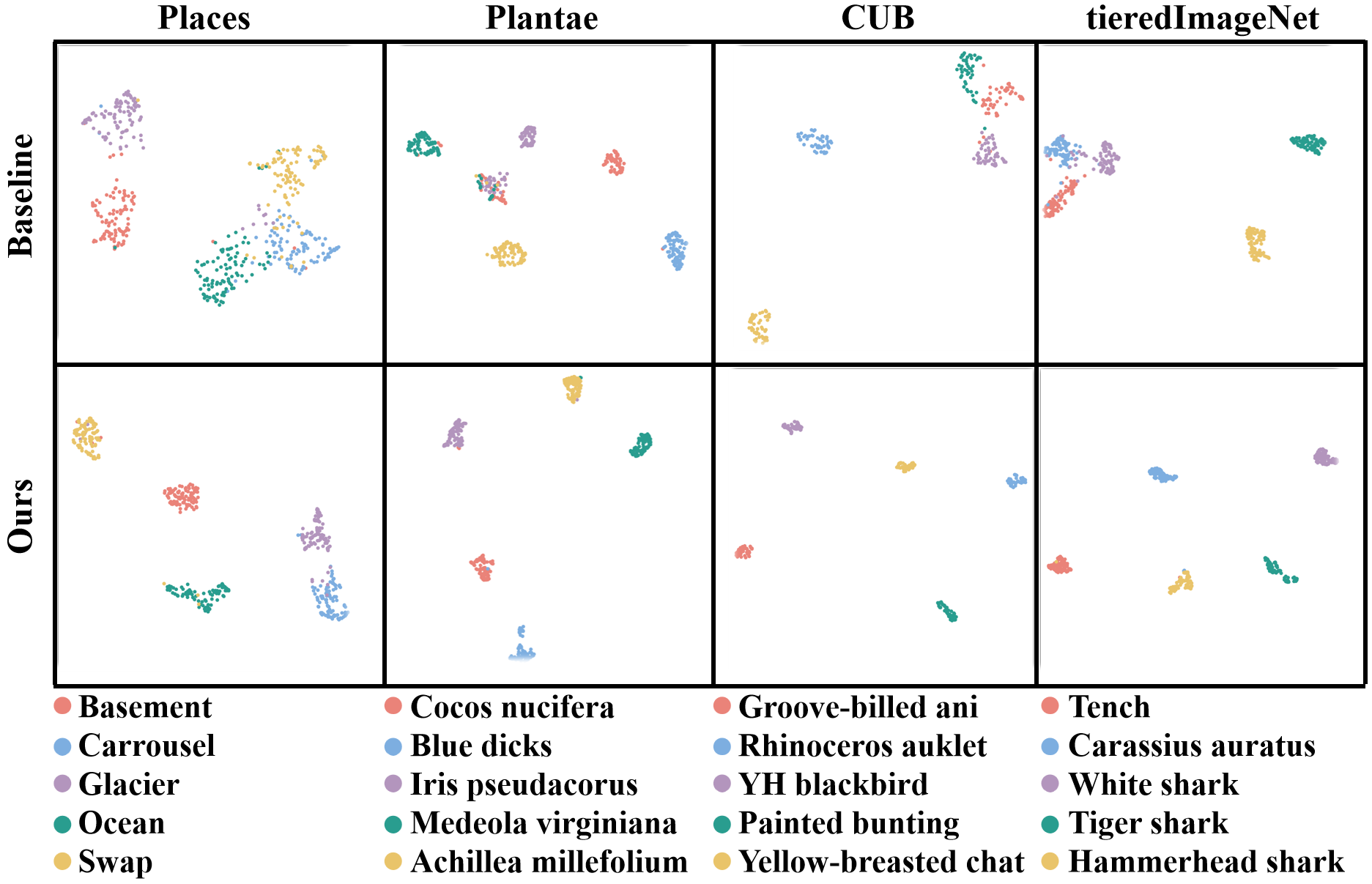}
\caption{UMAP visualization.}
\label{umap}
\end{figure*}
\subsection{A3. Statistical Analysis of the Dynamic Factor}
% 动态因子的平均值和方差

In order to evaluate the adaptability of the AUCA module on different datasets, we calculate the mean and variance of \(\lambda\). Specifically, we performed 1,000 tests per dataset to ensure reliable statistical results.
As shown in Table \ref{tab:lambda}, there are some differences in the mean values of \(\lambda \) on different domain datasets, which indicates that the AUCA module is able to adaptively generate uncertain classes according to the changes in the feature distributions, thus better adapting to different datasets. 
Moreover, the relatively small variances suggest that \(\lambda \) values remain stable within each dataset, demonstrating the robustness of the AUCA module in learning a consistent adaptation strategy.

\subsection{A4. UMAP Visualization}

Figure~\ref{umap} presents 2D UMAP visualizations of MPA features on five target-domain datasets. 
For each dataset, compared with the baseline, 
MPA clearly separates the category boundaries, demonstrating the effectiveness of MPA. 

\subsection{A5. Feature Visualization}
The results show that MPA effectively highlights target objects and captures more comprehensive semantic information, improving generalization. As shown in Fig. \ref{cam}, MPA performs well on both single-domain and cross-domain datasets, highlighting its semantic understanding and robustness to data distribution shifts.

\subsection{A6. Multimodal Information Richness}

To highlight the comprehensive multimodal information captured by our method, 
we compare MPA with existing baseline methods in terms of the types of information utilized: 
raw images, multi-view images, class names, attributes, and multi-variant semantics. 

\begin{table}[h]
\centering

\begin{tabular}{lccccc}
\toprule
Method       & Img & MVI & Cls & Attr & MVS \\
\midrule

SP-CLIP      & $\checkmark$ & $\times$           & $\checkmark$ & $\times$    & $\times$                  \\
CAML         & $\checkmark$ & $\times$           & $\checkmark$ & $\times$    & $\times$                  \\
SEVPro       & $\checkmark$ & $\times$           & $\times$     & $\times$    & $\times$                  \\
SemFew-Trans & $\checkmark$ & $\times$           & $\checkmark$ & $\checkmark$ & $\times$                 \\
SPM          & $\checkmark$ & $\times$           & $\checkmark$ & $\times$    & $\times$                  \\
ECER-FSL     & $\checkmark$ & $\times$           & $\times$     & $\checkmark$ & $\times$                 \\
MLVLM        & $\checkmark$ & $\times$           & $\checkmark$ & $\checkmark$ & $\times$                 \\
MPA (Ours)   & $\checkmark$ & $\checkmark$       & $\checkmark$ & $\checkmark$ & $\checkmark$            \\
\bottomrule
\end{tabular}
\caption{Comparison of the multimodal information richness across different methods. MVI represents multi-view images, and MVS represents multi-variant semantics.}
% \caption{Comparison of multimodal information richness across different inputs: Raw image(Img), Multi-View images (MVI), class names(Cls), attributes(Attr) and Multi-Variant Semantics(MVS).}
\label{tab:multimodal_richness}
\end{table}

As shown in Table~\ref{tab:multimodal_richness}, MPA leverages five types of information, yielding richer multimodal representations that improve prototype quality and few-shot performance.

\subsection{A7. Efficiency Analysis}
We evaluate computational efficiency on EuroSAT 5-way 1-shot. Table~\ref{tab:efficiency} shows memory (GB), runtime (s), and accuracy (\%) for different module combinations. LMSE and HMA slightly increase cost while boosting performance.

\begin{table}[h]
\centering

\begin{tabular}{cccccc}
\toprule
LMSE & HMA & Mem. (G) & Time (s) & Acc. (\%) \\
\midrule
$\times$ & $\times$ & 3.32 & 0.050 & 76.08 \\
$\checkmark$ & $\times$ & 3.36 & 0.058 & 83.03 \\
$\times$ & $\checkmark$ & 3.42 & 0.067 & 79.44 \\
\bottomrule
\end{tabular}
\caption{Efficiency comparison of LMSE and HMA on EuroSAT 5-way 1-shot.}
\label{tab:efficiency}
\end{table}
These results demonstrate that LMSE and HMA maintain high computational efficiency, with less than 3\% increase in memory and under 0.02,s additional test time per image.
The trade-off between accuracy and cost remains highly favorable, confirming that the proposed modules are lightweight and practical for few-shot learning tasks.

\end{document}